\documentclass[10pt,twocolumn,letterpaper]{article}

\usepackage{wacv}
\usepackage{times}
\usepackage{epsfig}
\usepackage{graphicx}
\usepackage{amsmath}
\usepackage{amssymb}

\usepackage[utf8]{inputenc} 
\usepackage[T1]{fontenc}    
\usepackage{hyperref}       
\usepackage{url}            
\usepackage{booktabs}       
\usepackage{amsfonts}       
\usepackage{nicefrac}       
\usepackage{microtype}      
\usepackage{algorithm}
\usepackage{algorithmic}
\usepackage{amssymb}
\usepackage{amsmath}
\usepackage{graphicx}
\usepackage{subfigure}
\usepackage{float}
\usepackage{stfloats}

\wacvfinalcopy 

\def\wacvPaperID{968} 

\ifwacvfinal\fi
\begin{document}

\title{Distributed Generative Adversarial Network}

\author{Xiaoyu Wang
{\tt\small wxystudio@xjtu.stu.edu.cn}
\and
Ye Deng
{\tt\small yedeng@stu.xjtu.edu.cn}
\and
jinjun wang
{\tt\small jinjun@mail.xjtu.edu.cn}
}

\maketitle
\ifwacvfinal\thispagestyle{empty}\fi

\begin{abstract}
   Recently the Generative Adversarial Network has become a hot topic. Considering the application of GAN in multi-user environment, we propose Distributed-GAN. It enables multiple users to train with their own data locally and generates more diverse samples. Users don't need to share data with each other to avoid the leakage of privacy. In recent years, commercial companies have launched cloud platforms based on artificial intelligence to provide model for users who lack computing power. We hope our work can inspire these companies to provide more powerful AI services.
\end{abstract}

\section{Introduction}

\par GAN is an unsupervised learning method that can automatically generate real data samples. Compared with other generative models, such as variational autocoder(VAE)\cite{kingma2013auto}, it does not work with an explicit density function and generate better samples. Therefore, GAN is widely used in transfer learning, data amplification, etc.

\par In recent years, many commercial companies have launched platforms based on deep learning method to provide users with AI services, such as AutoML. Users upload their data to the platforms and train a neural network. But this may lead to users' sensitive data leakage. \cite{shokri2015privacy}Reza Shokri proposed a distributed learning method, with which multiple users maintain their own data locally and jointly train a more powerful model. Google proposed a federate learning method based on this work and applied it to an application scenario. Users can train their own data on the mobile phone chip and communicate with the cloud platform to jointly train a more general model.

\par Inspired by this work, we propose Distributed-GAN. Considering the following scenario: in scientific research field, people do data augmentation to train more powerful model, and want to share data with partners. But for privacy consideration, we often avoid exposing our real data to each other. Our Distributed-GAN can solve the problem. It enables multiple users to participate in training with their real data locally, generate augmented data and share with each others. 

\par Our Distributed-GAN includes the following three types: The first is based on the distributed learning and federate learning method. When the platform applies distributed learning algorithm to train a model, we regard it as discriminator and use its output to train a generator, eventually get more data.

\par The second method combines the outputs of multiple discriminators to update one generator. We prove through experiment that this method has more remarkable effect when the users' data domains are similar.

\par The third method is to train one generator against multiple discriminators. We make a summary of the above three distributed methods, and find they have remarkable effect on reducing training time. We hope our work can enlighten the commercial companies to serve customers better AI services.

\par In summary, our contributions are as follows:

\par 1.In the first approach, we make use of the distributed learning method and extend it to Distributed-GAN, Which can not only train a discriminative model but also obtain augmented samples without exposing users' sensitive data.

\par 2.In the second approach, we propose the weighted-output algorithm and apply it to Distributed-GAN and find it is effective when users' data domain are similar. 

\par 3.In the third approach, we propose a naive Distributed-GAN, which directly train one generator against multiple discriminators, which show bright prospect in multi-user training enviroment. We show in experiment that all the above three methods can remarkably reduce training time. 

\section{related work}

\par In the privacy-preserving research field, \cite{shokri2015privacy}Reza Shokri proposed distributed learning algorithm to protect users' sensitive data. After that, Google proposed federate learning, which enables users to train data on mobile phone chip and jointly train a more powerful model with protecting users' privacy. 

\par \cite{goodfellow2014generative}Ian Goodfellow proposed GAN, which can automatically learn features from data and generate real data. \cite{arjovsky2017wasserstein}Martin Arjovsky proposed W-GAN and theoretically analyzed the reason why model collapse appears in the training process. \cite{radford2015unsupervised}Alec Radford proposed DC-GAN, replacing the fully connected layer with the convolutional layer, so that can obtain better result. \cite{mirza2014conditional}Mehdi Mirza proposed c-GAN in order to generate labeled sample. 

\par There are many types of combining many generators or discriminators for some tasks. CycleGAN\cite{zhu2017unpaired}, DiscoGAN\cite{kim2017learning} and DualGAN\cite{yi2017dualgan} are proposed to domain adaptation with combining two GAN architectures. \cite{liu2016coupled}Ming-Yu Liu proposed Couple-GAN with sharing some parameters of two generators. \cite{ghosh2018multi}Arnab Ghosh proposed MAD-GAN to train one discriminator against multiple generators. \cite{durugkar2016generative}Ishan Durugkar proposed MA-GAN to train one generator against multiple discriminators, but we emphasize that our training method is different from them and we focus on a a unique scenario, where users want to share data augmentation in distributed training situation without leaking privacy.

\begin{figure*}[hb]
\centering
\includegraphics[width=6in]{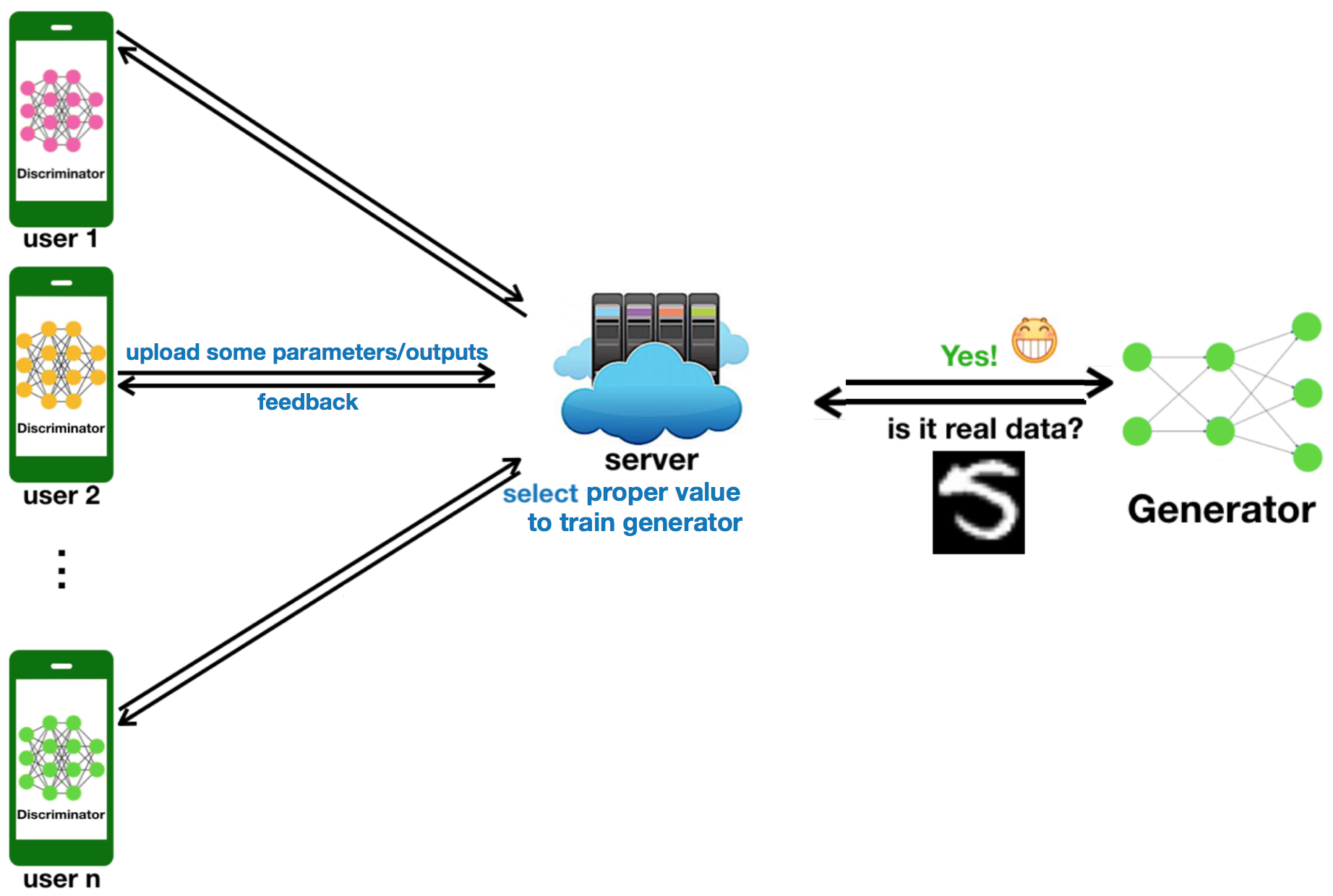} 
\caption{The first method of Distributed-GAN. N users train local discriminitor with distributed learning. We establish a neural network as server and update it with the weight uploaded by users, then train a generator against the neural network on platform.}
\end{figure*}

\section{Background}

\subsection{distributed learning}

Due to the potential hazard of leaking privacy in AI service platform, Reza Shokri proposed a distributed training approach. There are two advantages of this training method:
\par 1. Users can train a model without uploading data to the platform. 
\par 2. it enables each user to jointly train a more powerful and ubiquitous model, making full use of everyone's data.

\par First, all users need to agree on one neural network architecture, including the inner parameters and API. In the first training section, $n$ users first train their own neural networks with their own data, calculate the $LOSS$ function(for example, cross entropy), then get the partial derivative of each weight, finally obtain the gradient of each neuron. Each user uploads a portion of their gradients to the platform. 

\par After the platform collects the gradients of the neurons sent by each user, there are two strategies of selecting proper gradients. The first is platform randomly selects a fixed portion of gradients of each neuron as global parameter. The second is the platform selects some gradients bigger than a threshold as global parameter, and users update local model with the global parameter. 

\par Receiving the global gradient sent by the platform, each user updates their own weights with the obtained gradient, thereby obtaining a new neural network after one training epoch.

\par The above is the details of one training epoch. After that, each epoch is like the first epoch, continuously improves the accuracy of the training model, finally obtains the user's local model. In this scenario, the model obtained by the users has more powerful capability. For the MNIST dataset as example, if user 1 has numbers 0-8 without 9, and user 2 has numbers 0-9. After the training process, the final obtained model can identify 0-9.

\subsection{GAN}

\par GAN consists of a generator and a discriminator, the goal of generator is to generate samples similar to the real data, and the discriminator determines whether the generated data is true or fake. During the training, the generator and discriminator are designed to optimize the following formula:

$$\min\limits_{\theta_G} \max\limits_{\theta_D} \sum\limits_{i=1}^{n_+}logf(x_i;\theta_D) + 
\sum\limits_{j=1}^{n_-}log(1-f(g(z_j;\theta_G);\theta_D))$$

where $x_i$ and $x_j$ are relatively real data and fake data. $f$ is Discriminator and $g$ is Generator. $\theta_D$ and $\theta_G$ are relatively the parameter of two networks. 

The two compete against each other until the discriminator cannot discern the data generated by the generator, and we assume that the generator generates real data.

\section{Distributed-GAN}
\par Our Distributed-GAN is applied to the following scenario: GAN is widely applied to the data augmentation. We assume there are multiple users want to share their data to train a more powerful model, but do not want to expose their real data. So aiming at this problem, we propose three kinds of Distributed-GAN and hope to help commercial companies better serve users.

\subsection{distributed-GAN:the first approach}
\par Our first Distributed-GAN is directly inspired by the distributed learning algorithm. When the distributed learning platform works, we establish a neural network  as server and update it with the weight uploaded by users. At the same time, we build a generator on the platform. We propose an algorithm to calculate the output of this neural network. Then the output is used to train generator. The detail is shown in algorithm 1 and figure 1.

\begin{algorithm}
\caption{The first approach of Distributed-GAN. $w_i$ is the weight of user i. $w_s$ is the weight of server. $\theta _d$ is the parameter of generator, G is generator, D is discriminator.}
\begin{algorithmic}[1]
\FOR {j=0 to n epochs}         
    \STATE {users train local model using their own data}
    \STATE {user $i$ upload $\Delta w_i$ to server}
    \STATE {server selects the biggest $\Delta w_i$ as max($\Delta w_i$)}
    \STATE {server updates the server model parameter $w_s$ using max($\Delta w_i$)}
    \ENDFOR
\WHILE {the server model is updated}
    \STATE {producing fake\_images = G(z)} 
    \STATE {outputs = D(fake\_images)} 
    \STATE {using g\_loss = criterion(outputs, real\_labels) to train generator} 
\ENDWHILE
\end{algorithmic}
\end{algorithm}

\subsection{distributed-GAN:the second approach}
\par Ian Goodfellow proposed to replace (1-D (G)) with D (G), so as to make the GAN better converge. Here, we propose to use the average of two discriminators output to train the generator. We find that the generator trained in this way perform better when users' data domain are similar. The detail is shown in algorithm 2.

\begin{algorithm}
\caption{The second approach of Distributed-GAN. $G$ is generator, $D$ is discriminator. $z$ is the random noise for generator.}
\begin{algorithmic}[1]
\FOR {j=0 to n epochs}         
    \STATE {users train local model $D_i$ using their own data}
    \STATE {producing fake\_images = G(z)} 
    \STATE {outputs=($D_1$(fake\_images) + $D_2$(fake\_images)) $\backslash$ 2} 
    \STATE {using g\_loss = criterion(outputs, real\_labels) to train generator}
\ENDFOR
\end{algorithmic}
\end{algorithm}

\begin{algorithm}
\caption{The third approach of Distributed-GAN. G is generator, D is discriminator.}
\begin{algorithmic}[1]
\FOR {i=0 to n epochs}      
    \FOR {j=0 to m users}
        \STATE {$user_j$ train local model $D_j$ using their own data}
        \STATE {producing fake\_images = G(z)} 
        \STATE {outputs=$D_j$(fake\_images)} 
        \STATE {using g\_loss = criterion(outputs, real\_labels) to train generator}
    \ENDFOR
\ENDFOR
\end{algorithmic}
\end{algorithm}

\subsection{distributed-GAN:the third approach}
\par The third Distributed-GAN trains one generator against two discriminators simultaneously in one epoch. In original GAN, generated data is fed into discriminator to calculate $LOSS$ of generator. we propose that in one epoch we train the generator with the feedback of first discriminator and then with the second discriminator. We found that in this way the generator can generate multiple users' data. The detail is shown in algorithm 3.


\begin{figure*}[ht]
\centering
\includegraphics[width=6in,height=0.5in]{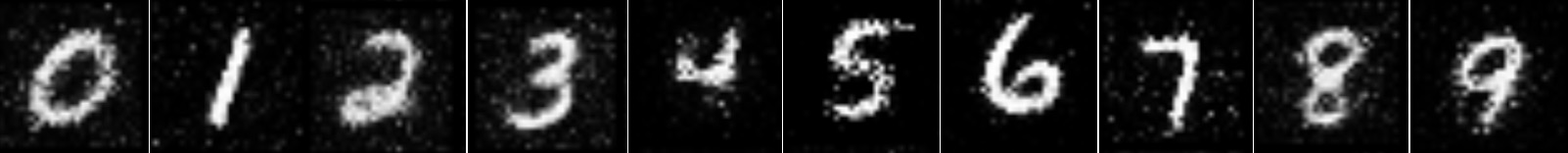} 
\caption{the first method for MNIST}
\end{figure*}

\begin{figure*}[ht]
\centering
\includegraphics[width=6in,height=0.5in]{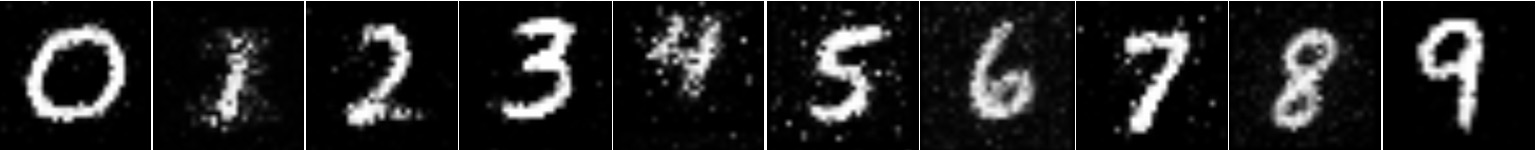} 
\caption{the second method for MNIST}
\end{figure*}

\begin{figure*}[hb]
\centering
\includegraphics[width=6in,height=1in]{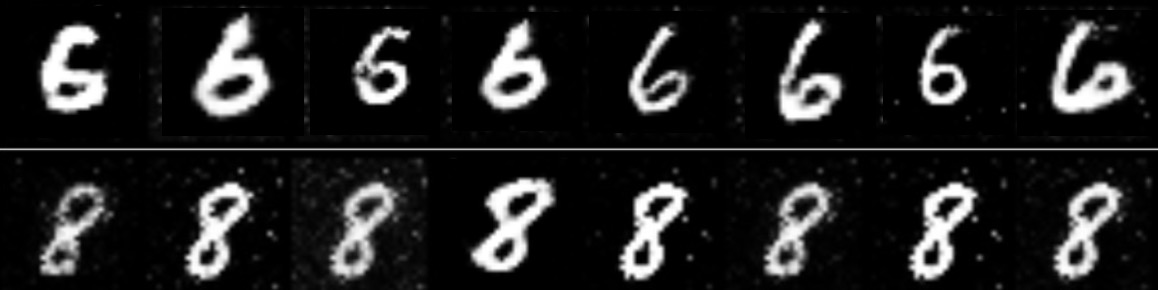} 
\caption{Train 6 an 8 with the second method. We can see obviously the result on 6 and 8 is better than the result on 4 and 7, because we find the domain of 6 and 8 is related(the manifold of two number pictures are similar.)}
\end{figure*}

\begin{figure*}[hb]
\centering
\includegraphics[width=6in,height=0.5in]{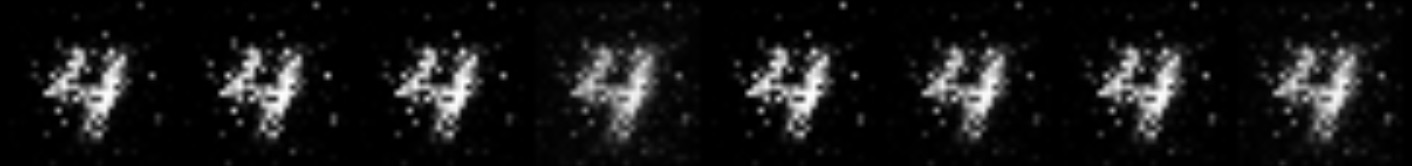} 
\caption{Train 4 an 7 with the second method. We can see the result on 4 and 7 is bad, because we propose the domain of 4 and 7 is far away(the manifold of two number pictures are unsimilar.) We even find that there is no 7 picture generated, it's because the domain we want to generate is too separate. We emphasize that the second method can only used when data domain is related.}
\end{figure*}

\begin{figure*}[ht]
\centering
\includegraphics[width=6in,height=0.5in]{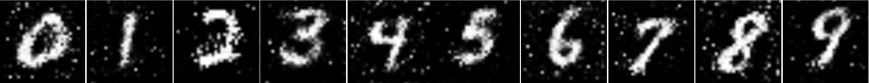} 
\caption{result of the third method training MNIST dataset. We split the 0-9 number for two users, one has 0-4 and another has 5-9. We find our Distributed-GAN can generate the whole 0-9 number without sharing users' data.}
\end{figure*}

\begin{figure*}[ht]
\centering
\includegraphics[width=6in,height=0.5in]{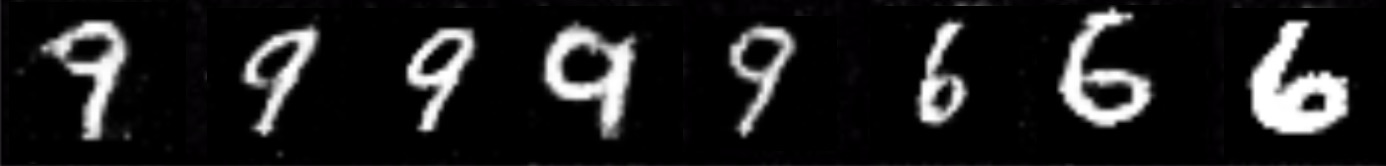} 
\caption{the third method for MNIST with 6 and 9. One user has only 6 and the other has 9, they jointly train Distributed-GAN model to obtain augmented data. The result shows our method can generate 6 and 9 without any data shared in two users.}
\end{figure*}

\section{Experiment}

\subsection{Experiment setup}
\par We train our Distributed-GAN with pytorch in MNIST dataset, CelebA dataset and LSUN dataset.
\par MNIST (Mixed National Institute of Standards and Technology database) is a computer vision dataset that contains 60,000 grayscale images of handwritten Numbers from 0 to 9, each of which contains 28x28 pixels. It also has 10000 test images.

\par CelebFaces Attributes Dataset (CelebA) is a large-scale face attributes dataset with more than 200K celebrity images, each with 40 attribute annotations. The images in this dataset cover large pose variations and background clutter. CelebA has large diversities, quantities, and rich annotations.

\par LSUN dataset contains around one million labeled images for each of 10 scene categories and 20 object categories. Because the whole dataset is very large, we choose the bedroom$-$train and church$-$outdoor$-$train packages to build our model. All the code is written with pytorch.

\subsection{The first approach}

\subsubsection{Overall result}
We divide the MNIST into two parts and each user obtain one piece. Figure 2 is result of the first approach.

\par We divide CelebA and LSUN dataset into two parts and each user obtain one. Due to the limit space we put the result of CelebA  and LSUN dataset in section 7 and 8 respectively. 

\subsection{The second approach}

\subsubsection{Overall result}

\par We divide the MNIST into 0-9 and distribute each number to each user. Two users train their own number dataset jointly with the average-output algorithm. We show in figure 3 that we can generate each number with the second approach. 
\par Due to the limit space we put the result of CelebA  and LSUN dataset in section 7 and 8 respectively. 

\subsubsection{Difference of data domain}
\par When the numbers that two users have are different, the experiment result varies. For example, figure 4 is the result of training number 6 and 8, figure 5 is the result of training 4 and 7. Obviously the previous result is better. We propose that the reason is because the domain difference of data. So we suggest that when users' dataset are similar they can use this method to train a better model.

\begin{figure*}[hb]
\begin{minipage}[t]{0.3\linewidth}
\centering
\includegraphics[width=1.7in]{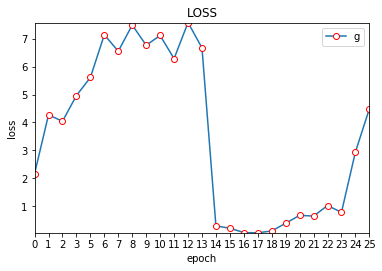}
\caption{first}
\label{fig:side:a}
\end{minipage}%
\begin{minipage}[t]{0.3\linewidth}
\centering
\includegraphics[width=1.7in]{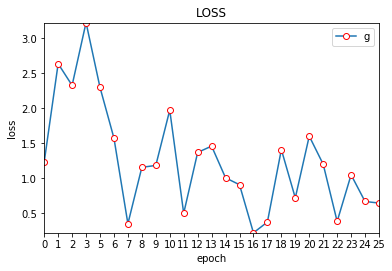}
\caption{second}
\label{fig:side:b}
\end{minipage}
\begin{minipage}[t]{0.3\linewidth}
\centering
\includegraphics[width=1.7in]{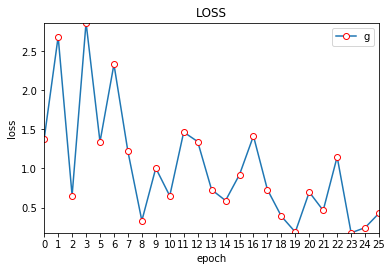}
\caption{third}
\label{fig:side:b}
\end{minipage}
\end{figure*}

\begin{figure*}[hb]
\begin{minipage}[t]{0.3\linewidth}
\centering
\includegraphics[width=1.7in]{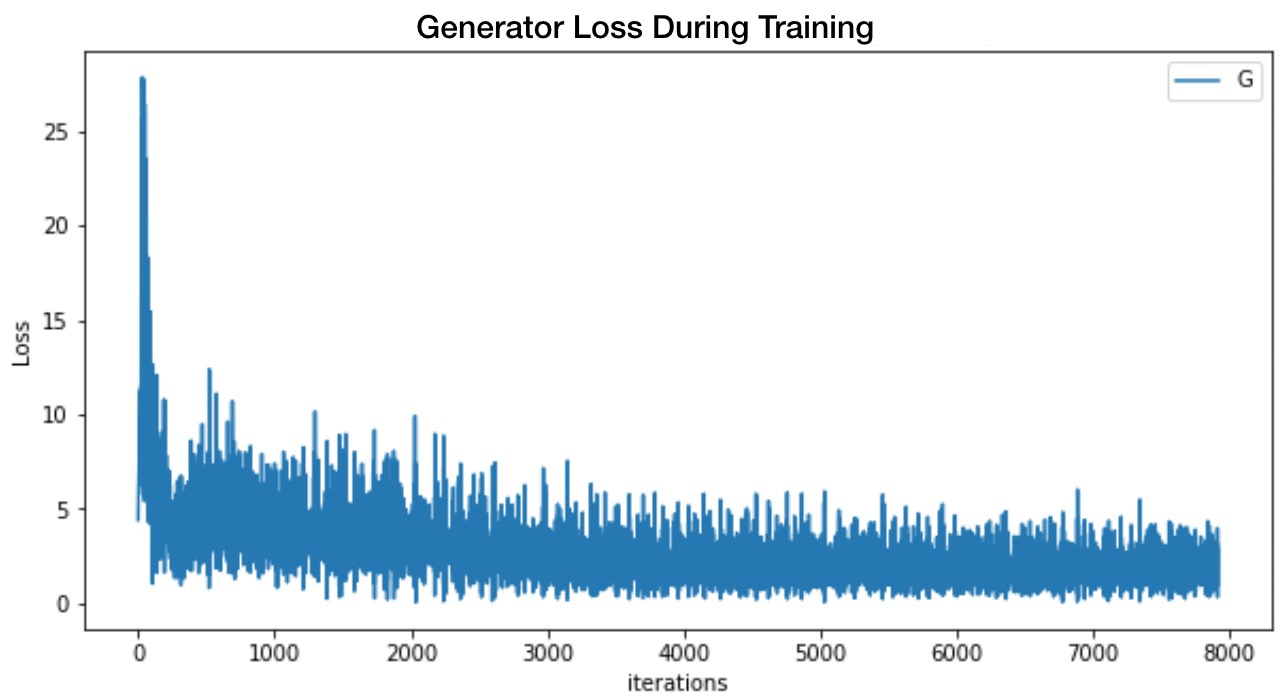}
\caption{first}
\label{fig:side:a}
\end{minipage}%
\begin{minipage}[t]{0.3\linewidth}
\centering
\includegraphics[width=1.7in]{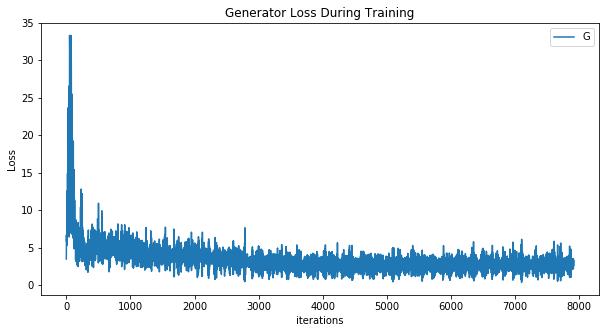}
\caption{second}
\label{fig:side:b}
\end{minipage}
\begin{minipage}[t]{0.3\linewidth}
\centering
\includegraphics[width=1.7in]{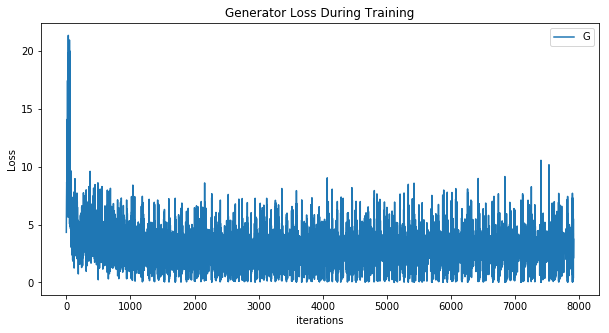}
\caption{third}
\label{fig:side:b}
\end{minipage}
\end{figure*}

\subsection{The third approach}

\subsubsection{Overall result}

We divide the MNIST dataset into two parts and each user obtain one piece, such as one has 0$-$4 and the other has 5$-$9, Figure 6 is the result of distributed training. Figure 7 is the result of two users train, when one obtains number 6 and another has 9.

\par Due to the limit space we put the result of CelebA  and LSUN dataset in section 7 and 8 respectively. 

\subsection{Time}
We propose that an obvious advantage of distributed training is saving time. Figure 14 shows the time consumed in training with our method and normal GAN. We can see there is a huge progress in time we spend.

\begin{figure}[H]
\begin{minipage}[t]{0.5\linewidth}
\centering
\includegraphics[width=1.5in]{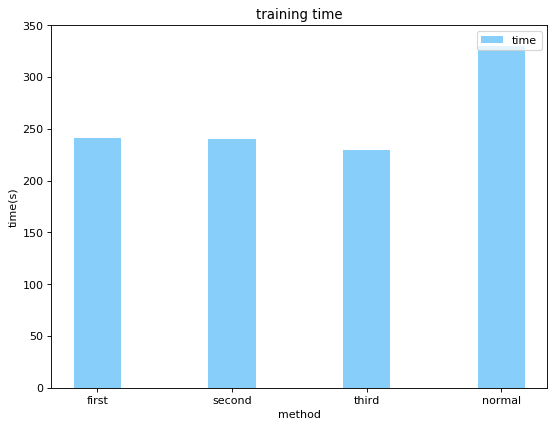}
\caption{Training time in MNIST dataset. We can see obviously our method time consuming is less than traditional training method.}
\label{fig:side:a}
\end{minipage}%
\begin{minipage}[t]{0.5\linewidth}
\centering
\includegraphics[width=1.5in]{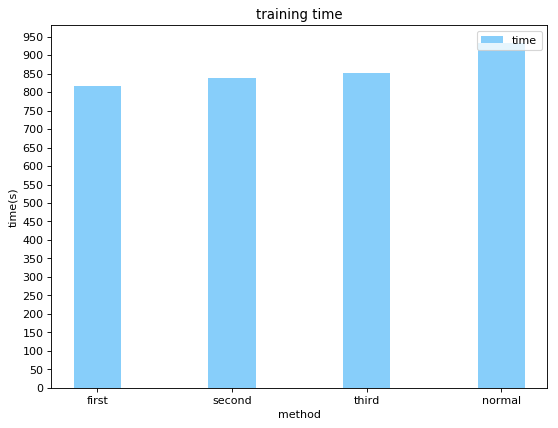}
\caption{Training time in CelebA dataset. Similar with above MNIST dataset, the time consuming is less than normal training.}
\label{fig:side:b}
\end{minipage}
\end{figure}

\subsection{LOSS}

\par We show the $LOSS$ of generator in training.There are the generator loss of three methods in training MNIST in figure 8-10. We can see a downtrend while training Distributed-GAN altough there are some instability. This proves our Distributed-GAN can be trained reliable.

\par There are the generator loss of three methods in training CelebA in figure 11-13.

\subsection{Large scale multi-user Distributed-GAN}
\par We implement our first and third method to large scale experiment. The result of multi-users training is put in section 9. For the first method, we split the dataset into many pieces and each user obtain one part. For the third method, we divide the dataset according to the label, and each user has one label. We emphasize it's also can split dataset as the first method.  

\section{SYSTEM ARCHITECTURE}

\begin{table}[H]
\caption{\large MNIST Discriminator}
\centering
\begin{tabular*}{5cm}{lll}
\hline
input$\rightarrow$(1)$\rightarrow$ ... $\rightarrow$(6)\\
\hline
(1). nn.Linear()\\
(2). nn.LeakyReLU()\\
(3). nn.Linear()\\
(4). nn.LeakyReLU()\\
(5). nn.Linear()\\
(6). nn.Sigmoid()\\
\hline
\end{tabular*}
\end{table}

\begin{table}[b]
\caption{\large MNIST Generator}
\centering
\begin{tabular*}{5cm}{lll}
\hline
input$\rightarrow$(1)$\rightarrow$ ... $\rightarrow$(6)\\
\hline
(1). nn.Linear()\\
(2). nn.ReLU()\\
(3). nn.Linear()\\
(4). nn.ReLU()\\
(5). nn.Linear()\\
(6). nn.tanh()\\
\hline
\end{tabular*}
\end{table}

\begin{table}[b]
\caption{\large CelebA Discriminator}
\centering
\begin{tabular*}{5cm}{lll}
\hline
input$\rightarrow$(1)$\rightarrow$ ... $\rightarrow$(13)\\
\hline
(1). nn.Conv2d()\\
(2). nn.LeakyReLU(max\_pool2d)\\
(3). nn.Conv2d()\\
(4). nn.BatchNorm2d()\\
(5). nn.LeakyReLU(max\_pool2d)\\
(6). nn.Conv2d()\\
(7). nn.BatchNorm2d()\\
(8). nn.LeakyReLU\\
(9). nn.Conv2d()\\
(10). nn.BatchNorm2d()\\
(11). nn.LeakyReLU\\
(12). nn.Conv2d()\\
(13). nn.sigmoid()\\
\hline
\end{tabular*}
\end{table}

\begin{table}[t]
\caption{\large CelebA Generator}
\centering
\begin{tabular*}{5cm}{lll}
\hline
input$\rightarrow$(1)$\rightarrow$ ... $\rightarrow$(14)\\
\hline
(1). nn.ConvTranspose2d()\\
(2). nn.BatchNorm2d()\\
(3). nn.ReLU()\\
(4). nn.ConvTranspose2d()\\
(5). nn.BatchNorm2d()\\
(6). nn.ReLU()\\
(7). nn.ConvTranspose2d()\\
(8). nn.BatchNorm2d\\
(9). nn.ReLU()\\
(10). nn.ConvTranspose2d()\\
(11). nn.BatchNorm2d\\
(12). nn.ReLU()\\
(13). nn.ConvTranspose2d()\\
(14). nn.Tanh()\\
\hline
\end{tabular*}
\end{table}

\section{Result of CelebA}

\begin{figure}[H]
\centering
\includegraphics[width=2.3in,height=3.8in]{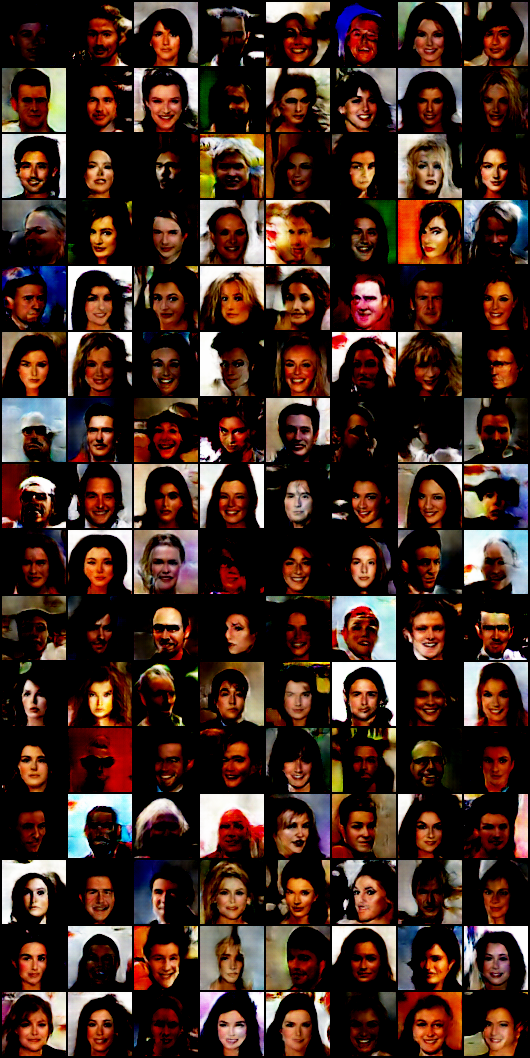} 
\caption{the first method for CelebA}
\end{figure}

\begin{figure}[H]
\centering
\includegraphics[width=2.3in,height=3.8in]{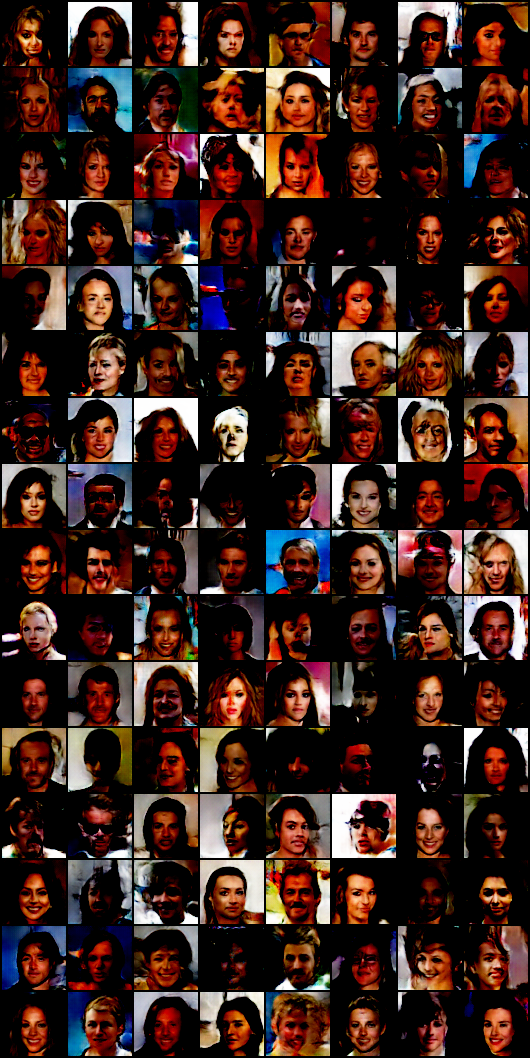} 
\caption{the second method for CelebA}
\end{figure}

\begin{figure}[H]
\centering
\includegraphics[width=2.3in,height=3.8in]{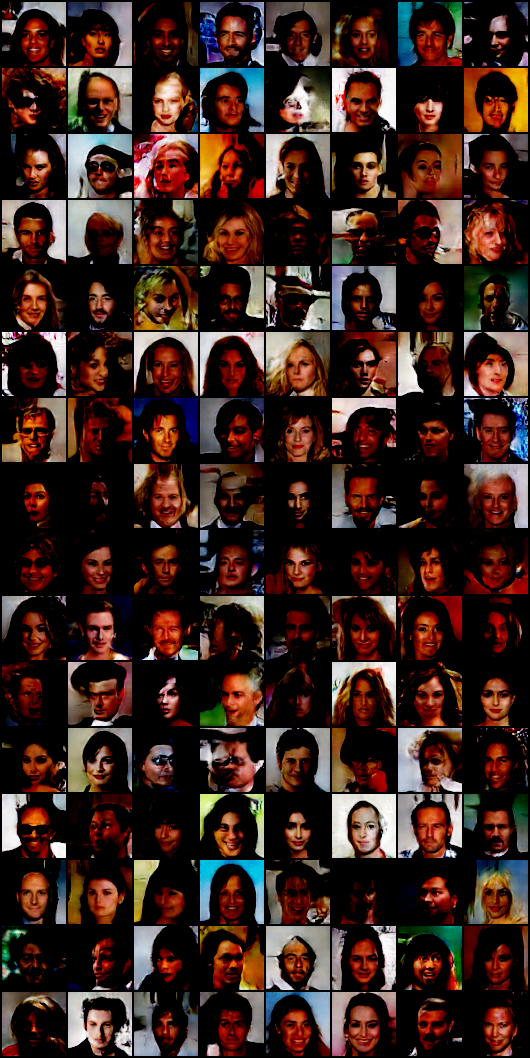} 
\caption{the third method for CelebA}
\end{figure}

\section{Result of LSUN}

\begin{figure}[H]
\centering
\includegraphics[width=2.3in,height=3.8in]{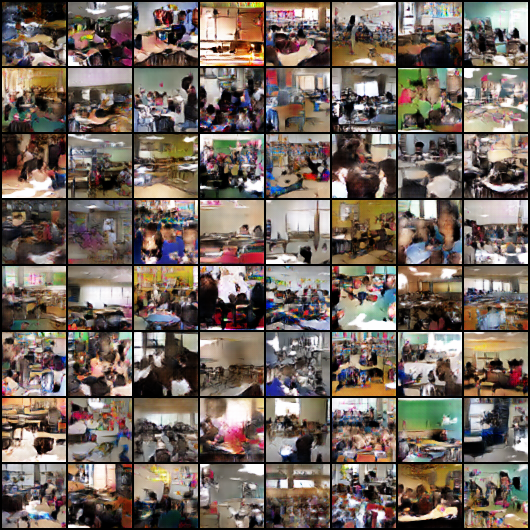} 
\caption{the first method for LSUN}
\end{figure}

\begin{figure}[H]
\centering
\includegraphics[width=2.3in,height=3.8in]{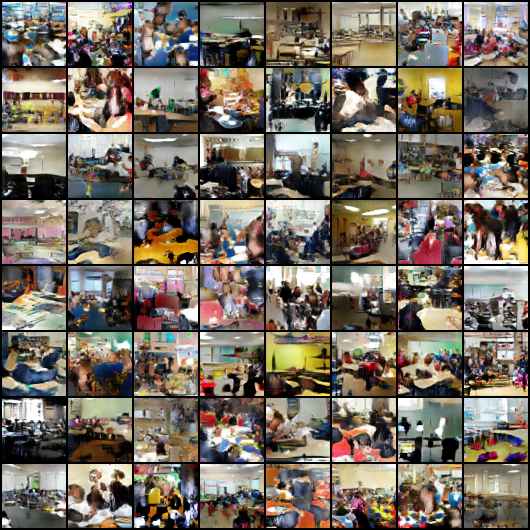} 
\caption{the second method for LSUN}
\end{figure}

\begin{figure}[H]
\centering
\includegraphics[width=2.3in,height=3.8in]{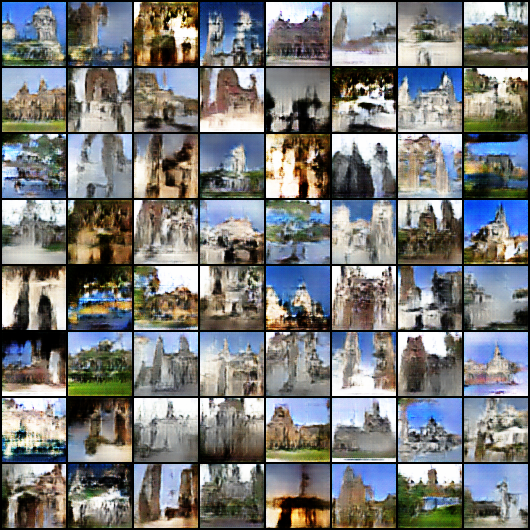} 
\caption{the third method for LSUN}
\end{figure}

\section{Result of multi-users training}

\begin{figure}[H]
\centering
\includegraphics[width=2.3in,height=3.8in]{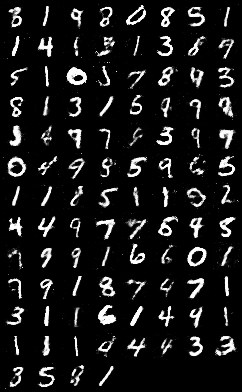} 
\caption{5 users train with MNIST}
\end{figure}

\begin{figure}[H]
\centering
\includegraphics[width=2.3in,height=3.8in]{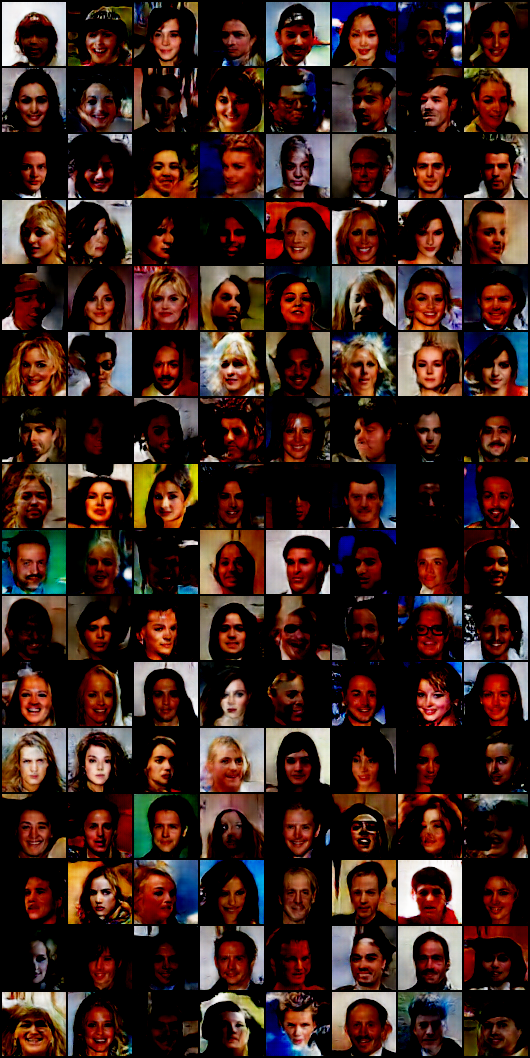} 
\caption{5 users train with CelebA}
\end{figure}

\section{Discussion}
\par We propose Distributed-GAN to certain need and scenario, but there are also some deficiency. Firstly we need to provide more rigorous derivation. The process of training and converging is unknown to us. Secondly we need to solve the notorious model collapse problem in GAN training. It also appears in distributed scenario. We hope to study these question in future work.

\section{Conclusion}
\par In view of the current hot topic, we propose three types of Distributed-GAN, with which users can jointly train a GAN to obtain augmented data without exposing their own sensitive data. We find that one kind of distributed-GAN has good effect when data domain is similar. In the future work, we need to improve the efficiency of information transmission in distributed system. At the same time, although our algorithm can protect users' privacy, there may be some ways to attack users' sensitive data.

{\small
\bibliographystyle{ieee}
\bibliography{egbib}
}

\end{document}